\documentclass[conference,a4paper]{APSIPA2025}
\usepackage{amsmath}
\usepackage{graphicx}
\usepackage{multirow}
\usepackage{threeparttable}
\usepackage[backend=biber,style=ieee,]{biblatex}
\usepackage{geometry}
\usepackage{fancyhdr}

\fancypagestyle{firststyle}{
  \fancyhf{}
  \fancyhead[C]{2025 Asia Pacific Signal and Information Processing Association Annual Summit and Conference (APSIPA ASC)}
}

\usepackage{nopageno}
\usepackage{nopageno} 
\usepackage{amsmath}  
\usepackage{amssymb}  
\usepackage{tikz}         
\usepackage{pgfplots}     

\usepackage{booktabs}

\usepackage{footmisc} 

\usepackage{footmisc} 
\renewcommand{\thefootnote}{\fnsymbol{footnote}} 

\newcommand{\restorenormalfootnotes}{%
    \renewcommand{\thefootnote}{\arabic{footnote}}%
}

\usepackage{amssymb}
\newcommand{\starnote}[1]{%
    \let\thefootnotemark=\relax
    \footnotetext{($\star$)\ #1}
}

\begin{document}

\title{NE-PADD: Leveraging Named Entity Knowledge for Robust Partial Audio
Deepfake Detection via Attention Aggregation}

\author{
\authorblockN{
Huhong Xian\authorrefmark{1},
Rui Liu\authorrefmark{1}\authorrefmark{2},
Berrak Sisman\authorrefmark{3} and
Haizhou Li\authorrefmark{4}
}


\authorblockA{
\authorrefmark{1}
Inner Mongolia University, Hohhot, China \\
E-mail: xianhuhong@163.com; liurui\_imu@163.com
}


\authorblockA{
\authorrefmark{3}
Center for Language and Speech Processing (CLSP), Johns Hopkins University, USA \\
E-mail: sisman@jhu.edu}

\authorblockA{
\authorrefmark{4}
School of Artificial Intelligence, The Chinese University of Hong Kong, Shenzhen, China \\
E-mail: haizhouli@cuhk.edu.cn }

}

\maketitle
\thispagestyle{firststyle}
\pagestyle{fancy}

\begin{abstract}
  Different from traditional sentence-level audio deepfake detection (ADD), partial audio deepfake detection (PADD) requires frame-level positioning of the location of fake speech. While some progress has been made in this area, leveraging semantic information from audio, especially named entities, remains an underexplored aspect. To this end, we propose a novel method, NE-PADD, which leverages named entity knowledge for robust partial audio deepfake detection through attention aggregation. NE-PADD consists of two parallel branches: Speech Named Entity Recognition (SpeechNER) and PADD. Specifically, we introduce two attention aggregation mechanisms to help PADD models better understand named entity knowledge, thus achieving more robust PADD performance. (1) Attention Fusion (AF) combines attention from SpeechNER and PADD for more accurate weights. (2) Attention Transfer (AT) uses an auxiliary loss from their attention distributions to guide PADD in learning named entity semantics. We construct an appropriative PartialSpoof-NER dataset based on the existing PartialSpoof dataset and conduct a detailed comparative analysis of the two attention aggregation methods. Experiments demonstrate that our method exceeds all advanced baselines and demonstrates the effectiveness of fusing named entity knowledge during PADD. The code is available at https://github.com/AI-S2-Lab/NE-PADD.
\end{abstract}

\section{Introduction}

\footnotetext{\authorrefmark{2} Corresponding author.}
\restorenormalfootnotes
In recent years, deep learning has catalyzed remarkable advancements in audio spoofing technologies, particularly in the domains of text-to-speech (TTS) \cite{wu2025diffcss,han2025stable,kang2023grad} and voice conversion (VC) \cite{li2023freevc,chan2022speechsplit,chen2021againvc,tang2022avqvc}. By leveraging generative models such as CosyVoice \footnote{https://github.com/FunAudioLLM/CosyVoice} and GPT-SoVITS \footnote{https://github.com/RVC-Boss/GPT-SoVITS}, TTS and VC systems now produce synthetic speech with near-human quality, rendering it nearly indistinguishable from genuine audio 
\cite{zuo2025enhancing,kong2020hifigan,van2016wavenet,shen2018naturaltts}. Partially spoofed audio refers to the insertion or splicing of synthesized speech segments into authentic audio, posing significant risks. Attackers manipulate small, specific units (such as words, characters, or even phonemes) to alter the meaning of sentences, thereby deceiving both machines and humans. Effectively detecting and mitigating such spoofed audio has become a critical and urgent challenge.


\begin{figure*}[t]
  \centering
  \includegraphics[width=\textwidth]{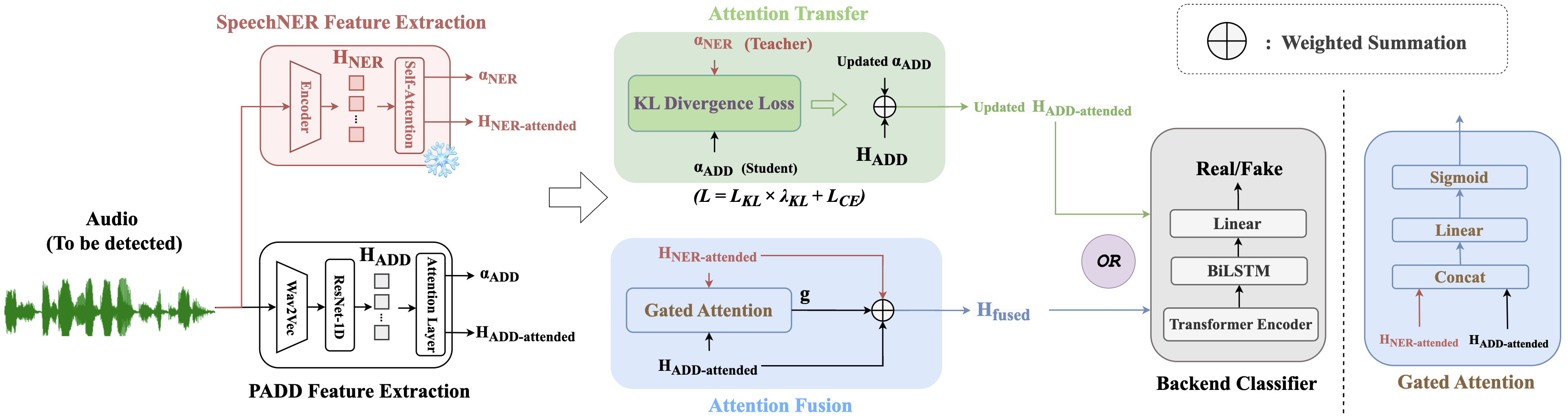} %
  \caption{The architecture of our proposed model. The model architecture includes two feature extraction modules on the left, with the proposed attention fusion (AF) and attention transfer (AT) methods in the center, and the backend classifier on the right.}
  \label{fig:architecture}
\end{figure*}

Currently, numerous methods and datasets have been developed for partial audio deepfake detection (PADD). RawNet2 \cite{tak2021end} processes raw audio using Sinc-Layers and residual blocks for efficient partial audio deepfake detection. AASIST \cite{jung2022aasist} combines a RawNet2-based encoder and RawGAT-ST for integrating temporal and spectral features using graph attention layers. Yi et al. \cite{yi2021half} created the first partially spoofed audio dataset, HAD, in which certain natural speech segments are replaced with synthesized speech segments that carry different semantic content. Zhang et al. \cite{zhang2021initial} introduced the PartialSpoof dataset, designed to provide segment-level labels for audio spoofing detection. Building on this dataset, they explored a multi-task learning framework capable of identifying spoofing occurrences at both the segment and utterance levels \cite{Zhang2021Arxiv}. Subsequently, they further extended this work, and Wav2vec2 \cite{baevski2020wav2vec} was adopted as a front-end feature extractor to enable simultaneous detection of both segment-level and utterance-level spoofing \cite{PartialSpoof2022}.

Although existing methods have made significant progress in PADD, the spoofed segments often appear in parts of the audio containing named entities. While previous PADD methods have demonstrated some effectiveness, they often overlook semantic information, particularly the role of named entities. In fact, named entities play a critical role in PADD. For example, partial deepfakes in everyday language frequently involve manipulating one or more named entities within a sentence. Consider the sentence: \textit{``CompanyX CEO heads a meeting in their Shenzhen office."}  Here, ``CompanyX" refers to an organization and ``Shenzhen" is a location. If any of these entities were replaced or altered, a partial deepfake could occur, which is often difficult to detect using traditional methods. Therefore, changes to named entities are crucial in PADD. Furthermore, some partial spoofing datasets (such as HAD) are specifically designed by replacing named entities, further emphasizing the central role of named entities in partial deepfake tasks. As such, exploring ways to leverage named entity information to improve PADD is an important research direction that warrants further investigation.

To address these challenges, we propose NE-PADD, a novel PADD framework that enhances detection by integrating semantic information from audio named entities using two attention mechanisms. We also introduce PartialSpoof-NER, a dataset based on PartialSpoof enriched with named entity annotations. Extensive experiments show that NE-PADD outperforms existing methods on key metrics such as Equal Error Rate, demonstrating its effectiveness and setting a new benchmark in partial audio deepfake detection. The main contributions are summarized as follows:
\begin{itemize}
    \item We propose a novel model called NE-PADD, which integrates the semantic information of named entities in audio into the deepfake detection model using an attention mechanism.

    \item We design an approach that integrates SpeechNER and PADD feature extraction modules to learn attention weights and extract intermediate features for spoof detection. We use Attention Fusion and Transfer to extract semantic information from speech named entities, ultimately improving the detection performance.

    \item We develop a new dataset called PartialSpoof-NER, based on the PartialSpoof dataset and enriched with named entity annotations, to provide semantic cues for spoofed audio detection.
\end{itemize}

\section{Proposed Method}

\thispagestyle{empty}

\subsection{Problem Statement and Overview}

In partially spoofed scenarios, spoofed audio segments are embedded within authentic speech. Our goal is to detect these spoofed segments at the frame-level. Given the large-scale self-supervised acoustic features:
\[
\mathbf{X} = \{f_1, f_2, \ldots, f_T\} \in \mathbb{R}^{D \times T},   \tag{1}
\]
where \(D\) and \(T\) represent the feature dimensions and the number of frames, respectively. The task is framed as a binary classification problem at the frame-level, where for each input feature \(\mathbf{X}\), the output is a sequence of frame-level labels:

\[
\mathbf{y} = \{y_1, y_2, \ldots, y_T\} \in \{0, 1\}^T. \tag{2}
\]

Here, \(y_t = 1\) indicates an authentic frame, while \(y_t = 0\) represents a spoofed frame. Since spoofed segments are often subtle and localized, achieving precise boundary detection is particularly challenging. 


In our proposed model, the attention fusion method generates gated attention weights through the Gated Attention module, which are then applied to the two distinct intermediate features, $\mathbf{H}_{\text{ADD-attended}}$ and $\mathbf{H}_{\text{NER-attended}}$, for weighted summation, producing a new fused feature $\mathbf{H}_{\text{fused}}$. The $\mathbf{H}_{\text{fused}}$ is subsequently passed to the frame-level classifier for the final detection of spoofed segments. The attention transfer method computes the Kullback–Leibler (KL) divergence between the two attention distributions, $\alpha_ \text{ADD}$ and $\alpha_ \text{NER}$, and incorporates it as an auxiliary loss, weighted by a specific factor into the main loss, effectively integrating the semantic information of named entities. The architecture of the proposed model is shown in Fig. 1.

\subsection{PADD Feature Extraction}

We utilize the pretrained Wav2Vec2.0 model, trained on the 960-hour Librispeech corpus, to extract frame-level features with a frame rate of 20ms \cite{cai2023waveform}. The features are further processed by ResNet-1D, which consists of multiple residual blocks. These blocks effectively capture temporal dependencies and local patterns in the audio, outputting embeddings ($\mathbf{H}_{\text{ADD}}$) that serve as input to the attention mechanisms.

Next, the $\mathbf{H}_{\text{ADD}}$ are passed through an attention layer to dynamically focus on relevant regions of the audio. The attention scores are calculated as:

\[
\alpha_{\text{ADD}} = \text{softmax} \left( \frac{\mathbf{Q}_{\text{ADD}} \mathbf{K}_{\text{ADD}}^\top}{\sqrt{D}} \right).  \tag{3}
\]

Here, \(\alpha_{\text{ADD}}\) represents the dynamic attention weights that highlight potentially spoofed segments in the audio. The attended embeddings (\(\mathbf{H}_{\text{ADD-attended}}\)) are computed as:

\[
\mathbf{H}_{\text{ADD-attended}} = \alpha_{\text{ADD}} \mathbf{V}_{\text{ADD}}.  \tag{4}
\]

The attention-modified embeddings (\(\mathbf{H}_{\text{ADD-attended}}\)) provide a focused representation of the audio input, emphasizing regions that are likely to contain spoofed segments.

\subsection{SpeechNER Feature Extraction}


The SpeechNER feature extraction module identifies named entities by adding special symbols before and after the named entities in the transcribed text of the input audio. Given the scarcity of research on Chinese speech named entities, our experiments are currently limited to English datasets. We draw inspiration from the work of Yadav et al. \cite{yadav2020end} on end-to-end named entity recognition from English speech, focusing on recognizing the three most frequent named entities (organization, person, and location) from English speech datasets. For this purpose, we use three special symbols (`\{', `\texttt{|}', `\$') to denote the start of different named entities, with `]' serving as a common symbol to mark the end of all named entities. As illustrated by the audio example proposed in the Introduction section of our paper: \textit{``CompanyX CEO heads a meeting in their Shenzhen office."}  The actual training text label appears as \textit{``\{CompanyX] CEO heads a meeting in their \$Shenzhen] office,"} where `\{' and `\$' denote the start of named entities, and `]' marks their end. This is treated as a sequence labeling task, where the input audio is transformed into log-spectrograms of power-normalized audio clips using a 20~ms Hamming window \cite{yadav2020end}, following the common practice in speech processing.

The SpeechNER feature extraction module first processes the named entity information from the audio input through an encoder. The encoder handles audio features, including convolutional and normalization layers, followed by multiple bidirectional long short-term memory (BiLSTM) layers to capture temporal dependencies in the audio data. This process generates named entity representations ($\mathbf{H}_{\text{NER}}$) suitable for further processing, encapsulating contextual information at each time step of the audio.  

Subsequently, a self-attention mechanism is applied to process these representations. The attention mechanism computes attention scores to generate entity-level embeddings. These attention-modified embeddings ($\mathbf{H}_{\text{NER-attended}}$) highlight the significance of named entities, providing crucial semantic information for the audio deepfake detection task.


\[
\alpha_{\text{NER}} = \text{softmax} \left( \frac{\mathbf{Q}_{\text{NER}} \mathbf{K}_{\text{NER}}^\top}{\sqrt{D}} \right) ,  \tag{5}
\]

\[
\mathbf{H}_{\text{NER-attended}} = \alpha_{\text{NER}} \mathbf{V}_{\text{NER}}.  \tag{6}
\]


\subsection{Attention Fusion}

The attention fusion mechanism merges $\mathbf{H}_{\text{ADD-attended}}$ and $\mathbf{H}_{\text{NER-attended}}$ through a gated attention strategy. Specifically, the fusion weight $g$ is computed as
\[
g = \sigma\left(W_g \left[ \mathbf{H}_{\text{ADD-attended}}, \mathbf{H}_{\text{NER-attended}} \right]\right),   \tag{7}
\]
where $\sigma$ denotes the Sigmoid function, and $[\cdot, \cdot]$ represents concatenation. Because of the Sigmoid activation, the fusion weight $g$ is constrained to the range $[0,1]$, where a value of $0$ indicates full reliance on $\mathbf{H}_{\text{NER-attended}}$, a value of $1$ indicates full reliance on $\mathbf{H}_{\text{ADD-attended}}$, and intermediate values represent a weighted combination of the two.

The fused features are then calculated as
\[
\mathbf{H}_{\text{fused}} = g \cdot \mathbf{H}_{\text{ADD-attended}} + (1 - g) \cdot \mathbf{H}_{\text{NER-attended}}.  \tag{8}
\]
which are subsequently fed into the backend classifier.




\subsection{Attention Transfer}

The attention transfer mechanism aligns \(\alpha_{\text{ADD}}\) with \(\alpha_{\text{NER}}\), enabling the teacher weights \(\alpha_{\text{NER}}\) to guide the student weights \(\alpha_{\text{ADD}}\) in learning semantic information. Specifically, this alignment is achieved by minimizing the KL divergence between the two distributions:

\[
\mathcal{L}_{\text{KL}} = \frac{1}{N} \sum_{i=1}^N \sum_{j=1}^T \alpha_{\text{NER}, j} \log \frac{\alpha_{\text{NER}, j}}{\alpha_{\text{ADD}, j}} .  \tag{9}
\]



Due to its asymmetry, KL divergence is employed in a teacher–student manner:
\(\alpha_{\text{NER}}\) serves as the teacher and \(\alpha_{\text{ADD}}\) as the student.
Minimizing \(KL(\alpha_{\text{NER}}\|\alpha_{\text{ADD}})\) ensures one-way
semantic knowledge transfer from SpeechNER to PADD.

The total loss combines the primary binary cross-entropy (BCE) loss with the transfer loss:

\[
\mathcal{L} = \mathcal{L}_{\text{CE}} + \lambda_{\text{KL}} \cdot \mathcal{L}_{\text{KL}}.   \tag{10}
\]

Here, \(\lambda_{\text{KL}}\) is a hyperparameter that balances the two components. This new loss guides the update of \(\alpha_{\text{ADD}}\), which is then used to compute attention on \(\mathbf{H}_{\text{ADD}}\), producing \(\mathbf{H}_{\text{ADD-attended}}\), the input to the backend classifier.


\subsection{Backend Classifier}
\thispagestyle{empty}

The embeddings are passed to the frame-level backend classifier for final prediction. To capture long-range global context within each frame, we employ multiple Transformer encoders. Subsequently, we apply a BiLSTM to further model the sequence embeddings generated by the Transformer encoders. Finally, a fully connected layer is used to predict the boundary probability for each frame.

\section{Experiments and Results}

\subsection{Data Preparation}

To rigorously validate the proposed method, we constructed the PartialSpoof-NER dataset, which incorporates named entity information. This dataset was derived from the PartialSpoof dataset by leveraging a multi-step process. Specifically, the audio data from the PartialSpoof dataset was first transcribed into text using the ASR model Whisper \cite{radford2023robust}. Subsequently, named entities were extracted from the transcriptions using Stanza \footnote{https://github.com/stanfordnlp/stanza}, a tool developed by the Stanford University research team. The resulting dataset, annotated with named entity information, was then curated as PartialSpoof-NER. 

Table I presents the statistics of the datasets used in our experiments. As shown in Table I, the Train dataset consists of 966 bona fide utterances and 8,789 fake utterances, with a total of 11,572 named entities. The Dev dataset contains 124 bona fide utterances and 1,057 fake utterances, with a total of 1,407 named entities. The Eval dataset includes 122 bona fide utterances and 1,126 fake utterances, encompassing 1,526 named entities in total. Each dataset contains, on average, 1 to 2 named entities per utterance.

\begin{table}[t!]
    \caption{The statistics of datasets. }
    \label{tab:ablation_results}
    \centering
    \begin{tabular}{ccccc}
        \toprule
        \textbf{Name}& Bona fide &Fake &All &Named Entities Count\\
        \midrule
        Train& 966& 8789&9755 &11572\\
 Dev&124& 1057&1181 &1407\\
        Eval& 122& 1126&1248 &1526\\
        \bottomrule
 & & & &\\
    \end{tabular}
\end{table}




\subsection{Experiment Setup}

The model is optimized using binary cross-entropy loss with the Adam optimizer over 100 epochs. Training is conducted with a mini-batch size of 16 and a learning rate of $10^{-4}$, with a Noam scheduler \cite{vaswani2017attention} applied for learning rate adjustment, including 1,600 warm-up steps. During training, the equal error rate (EER) is periodically evaluated on the adaptation set to monitor model performance.


In NE-PADD, the input audio is represented with dimensions \( L \times 1 \), where \( L \) denotes the audio length. The parameters of the SpeechNER feature extraction module are frozen. In the PADD feature extraction module, the input is processed with Wav2Vec, and the output is transformed to a size of \( T \times 768 \), where \( T \) represents the number of time stamps. The first 1D convolutional layer is configured with \( C(5, 2, 1) \), without bias, producing an output of \( T \times 512 \). This is followed by 12 residual blocks, each consisting of convolution and normalization layers with \( C(1, 0, 1) \), maintaining the output size at \( T \times 512 \). A second 1D convolutional layer with the structure \( C(1, 0, 1) \) further reduces the output size to \( T \times 128 \). 

The backend classifier uses a Transformer encoder with \( E(2, 4, 1024) \), which consists of 2 layers, 4 attention heads, and a feedforward network size of 1024, maintaining the output dimensions at \( T \times 128 \). Finally, a BiLSTM layer with 1 hidden layer of size 128 produces an output of \( T \times 128 \), followed by a ReLU activation function; ultimately, a 256-d fully-connected layer is used to predict the probability of each frame.

\begin{table}[t!]
    \caption{Comparison of EER (\%) results among different models, including our proposed NE-PADD framework (reported results represent NE-PADD-AF as the optimal implementation).}
    \label{tab:ablation_results}
    \centering
    \begin{tabular}{cc}
        \toprule
        \textbf{Model} & \textbf{EER} \\
        \midrule
        Single reso. \cite{PartialSpoof2022}& 26.51\\
 BAM \cite{zhong2024enhancing}&18.21\\
        TDL \cite{xie2024efficient} & 14.64\\
        WBD \cite{cai2023waveform}& 11.59\\
        NE-PADD& \textbf{7.89}\\
        \bottomrule
 &\\
    \end{tabular}
\end{table}

\begin{table}[t!]
    \caption{Comparison of experimental results after applying attention fusion (AF) and attention transfer (AT) to each baseline model. 
}
    \label{tab:ablation_results}
    \centering
    \begin{tabular}{cc}
        \toprule
        \textbf{Model} & \textbf{EER} \\
        \midrule
        BAM-AT& 14.84\\
 BAM-AF&10.61\\
        TDL-AT& 8.87\\
        TDL-AF& 9.28\\
        NE-PADD-AT& 9.48\\
        NE-PADD-AF& \textbf{7.89}\\
        \bottomrule
 &\\
    \end{tabular}
\end{table}

\subsection{Baselines}

To demonstrate the effectiveness of the proposed NE-PADD model, we compared it against four state-of-the-art PADD models: \textbf{1) Single reso.} \cite{PartialSpoof2022} innovates in architecture by introducing SSL models and training strategies using segment-level labels for multi-resolution training in the PS spoofing scenario. \textbf{2) BAM} \cite{zhong2024enhancing} utilizes boundary information as auxiliary attention for partially spoofed audio localization and develops a two-branch boundary module to exploit frame-level information for discriminative features. \textbf{3) TDL} \cite{xie2024efficient} combines an embedding similarity module and temporal convolution operation to effectively capture both feature and positional information. \textbf{4) WBD} \cite{cai2023waveform} proposes a frame-level boundary detection system using segment discontinuity and acoustic information.

\subsection{Metrics}
\thispagestyle{empty}


We use the Equal Error Rate (EER) to evaluate our model, a standard metric in binary classification tasks. Although our task involves frame-level partial spoof detection, each frame prediction can still be treated as a binary decision, making EER a relevant and informative measure. The EER corresponds to the point where the false positive rate equals the false negative rate, providing a balanced assessment of detection performance. A lower EER indicates higher overall accuracy in distinguishing between genuine and spoofed frames.

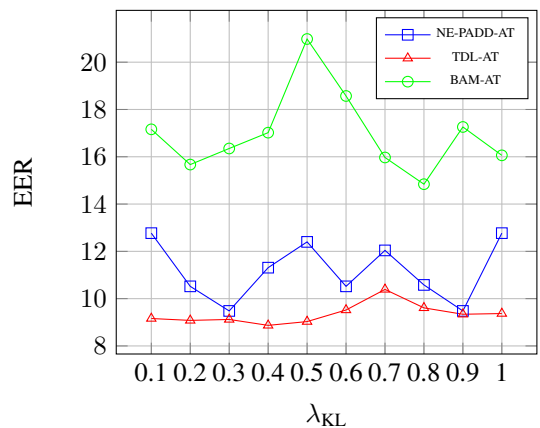
\begin{figure}[ht!]
    \centering
    \begin{tikzpicture}
        \begin{axis}[
            title={ },
            xlabel={\(\lambda_{\text{KL}}\)},
            ylabel={EER},
            xtick={0.1, 0.2, 0.3, 0.4, 0.5, 0.6, 0.7, 0.8, 0.9, 1},
            xticklabels={0.1, 0.2, 0.3, 0.4, 0.5, 0.6, 0.7, 0.8, 0.9, 1},
           ytick={8, 10, 12, 14, 16, 18, 20},
            width=0.8\linewidth,
            grid=major,
            legend style={font=\tiny},
        ]
        \addplot[
            color=blue,
            mark=square,
            ] coordinates {
            (0.1, 12.77) (0.2, 10.52) (0.3, 9.48) (0.4, 11.31) (0.5, 12.4) (0.6, 10.52) (0.7, 12.04) (0.8, 10.58) (0.9, 9.48) (1, 12.77)
        };
        \addlegendentry{NE-PADD-AT}

        \addplot[
            color=red,
            mark=triangle,
            ] coordinates {
            (0.1, 9.16) (0.2, 9.08) (0.3, 9.12) (0.4, 8.87) (0.5, 9.03) (0.6, 9.52) (0.7, 10.39) (0.8, 9.61) (0.9, 9.34) (1, 9.37)
        };
        \addlegendentry{TDL-AT}

        \addplot[
            color=green,
            mark=o,
            ] coordinates {
            (0.1, 17.16) (0.2, 15.67) (0.3, 16.35) (0.4, 17.02) (0.5, 20.98) (0.6, 18.57) (0.7, 15.97) (0.8, 14.84) (0.9, 17.26) (1, 16.06)
        };
        \addlegendentry{BAM-AT}

        \end{axis}
    \end{tikzpicture}
    \caption{EER Results for Different Values of \(\lambda_{\text{KL}}\)}
    \label{fig:eer_results}
\end{figure}

\subsection{Comparison with Existing Methods}


As shown in Table II, our attention fusion method achieved an EER of just 7.89\%, achieving the best performance on the PartialSpoof-NER dataset. It is worth clarifying that NE-PADD refers to our overall framework that leverages named entity semantic information to enhance partial audio deepfake detection. Within this framework, we explore two concrete implementation strategies: NE-PADD-AF and NE-PADD-AT. Both strategies follow the same core idea of exploiting speech-NER semantic weights, but differ in how the attention signals are integrated. Since NE-PADD-AF delivers the best empirical results, we report its performance as the representative implementation of NE-PADD when comparing with other baselines. By introducing these innovative attention fusion and attention transfer methods under the NE-PADD framework, our model achieves more precise capture of named entities’ semantic weights, leading to superior outcomes in the PADD tasks.

\subsection{Ablation study}


We conducted ablation experiments to evaluate the effectiveness of integrating semantic information in enhancing the performance of PADD models. By applying our proposed attention fusion and attention transfer methods to various models, we enabled them to learn semantics directly from audio signals. As shown in Table III, experimental results reveal that the BAM model achieved significant EER reductions of 3.37\% and 7.60\% when incorporating the attention transfer and attention fusion methods, respectively, compared to the baseline model. Similarly, the TDL model exhibited EER decreases of 5.77\% and 5.36\% through the implementation of attention transfer and attention fusion methods. These findings demonstrate that introducing semantic information via attention mechanisms substantially improves detection capabilities in PADD models. The results underscore the critical role of semantic features in identifying subtle audio manipulations and validate the efficacy of integrating such features into the detection framework. This research highlights the importance of leveraging semantic understanding for combating increasingly sophisticated audio forgery techniques. For clarity, when we denote variants such as BAM-AT and BAM-AF, or TDL-AT and TDL-AF, they refer to the baseline BAM (or TDL) model augmented with our AT or AF mechanisms, respectively. This naming convention is consistent with the NE-PADD framework described earlier.

For the attention transfer method, as shown in Fig. 2, we conducted multiple experiments on three models by varying the hyperparameter $\lambda_{\text{KL}}$ in steps of 0.1, ranging from 0.1 to 1. The results for attention transfer presented in Table III reflect the optimal hyperparameter values for each model.

\thispagestyle{empty}
\subsection{Finer-grained resolution experiment}

To investigate the impact of the number of spoofed segments in an audio file on the model's detection performance, we split the existing dataset by the number of spoofed segments (e.g., 1 forgery, 2 forgeries, 3 forgeries, up to 10 forgeries) and trained and tested the model on these subsets. As shown in Fig. 3, while the limited size of the dataset may prevent the results from fully reflecting the model’s generalization ability, we observed that as the number of spoofed segments increased, the model's accuracy in detecting forgeries improved, as evidenced by a lower EER. This indicates that the number of spoofed segments has a significant impact on model performance.

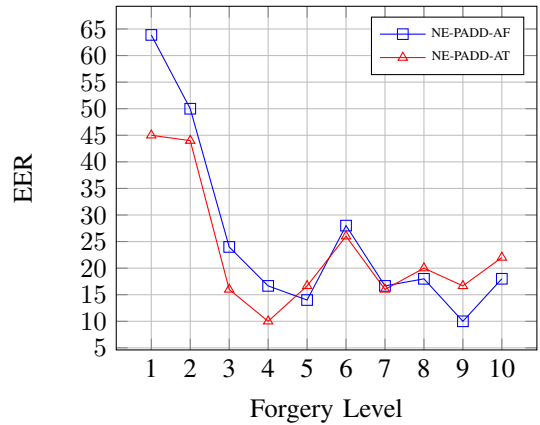
\begin{figure}[ht!]
    \centering
    \begin{tikzpicture}
        \begin{axis}[
            title={},
            xlabel={Forgery Level},
            ylabel={EER},  
            xtick={1,2,3,4,5,6,7,8,9,10},
            xticklabels={1, 2, 3, 4, 5, 6, 7, 8, 9, 10},
            legend pos=north east,
            grid=major,
            ytick={0,5,10,15,20,25,30,35,40,45,50,55,60,65},  
            width=0.8\linewidth,
            legend style={font=\tiny}
        ]
        \addplot[
            color=blue,
            mark=square,
            ] coordinates {
            (1,63.88) (2,49.99) (3,23.99) (4,16.66) (5,14.00) (6,28.00) (7,16.66) (8,17.99) (9,10.00) (10,18.00)
        };
        \addlegendentry{NE-PADD-AF}

        \addplot[
            color=red,
            mark=triangle,
            ] coordinates {
            (1,45.00) (2,43.99) (3,15.99) (4,10.00) (5,16.66) (6,26.00) (7,15.99) (8,20.00) (9,16.66) (10,22.00)
        };
        \addlegendentry{NE-PADD-AT}

        \end{axis}
    \end{tikzpicture}
    \caption{EER results for the dataset with different levels of segmentation forgery.}
    \label{fig:eer_results}
\end{figure}

\section{Conclusions}

In this paper, we propose a novel method for partial deepfake audio detection and introduce PartialSpoof-NER, a new dataset. Our model uses two attention mechanisms to fuse semantic features, improving detection of subtle audio manipulations. It achieves state-of-the-art performance on PartialSpoof-NER. Detecting very short spoofed segments remains challenging and will be a focus of future work. Due to limited data, the model has not yet been applied to Chinese–English partial forgery; future work will expand data resources to validate multilingual generalization.

\section{Acknowledgment}
The research by Rui Liu was funded by the Young Scientists Fund (No.~62206136), the General Program (No.~62476146) of the National Natural Science Foundation of China, the Young Elite Scientists Sponsorship Program by CAST (2024QNRC001), the Outstanding Youth Project of Inner Mongolia Natural Science Foundation (2025JQ011), Key R\&D and Achievement Transformation Program of Inner Mongolia Autonomous Region (2025YFHH0014) and the Central Government Fund for Promoting Local Scientific and Technological Development (2025ZY0143). The work of Huhong Xian was funded by the Research and Innovation Project for Graduate Students of Inner Mongolia University. The work by Berrak Sisman was supported by NSF CAREER award IIS-2338979. The work by Haizhou Li was supported by the Shenzhen Science and Technology Program (Shenzhen Key Laboratory, Grant No. ZDSYS20230626091302006), the Shenzhen Science and Technology Research Fund (Fundamental Research Key Project, Grant No. JCYJ20220818103001002), and the Program for Guangdong Introducing Innovative and Enterpreneurial Teams, Grant No. 2023ZT10X044.










\printbibliography

@inproceedings{shen2018naturaltts,
  author    = {J. Shen and R. Pang and R. J. Weiss and M. Schuster and N. Jaitly and Z. Yang and Z. Chen and Y. Zhang and Y. Wang and R. Skerry-Ryan and others},
  title     = {Natural TTS Synthesis by Conditioning Wavenet on Mel Spectrogram Predictions},
  booktitle = {IEEE International Conference on Acoustics, Speech and Signal Processing (ICASSP)},
  year      = {2018},
  pages     = {4779--4783}
}

@inproceedings{kong2020hifigan,
  author    = {J. Kong and J. Kim and J. Bae},
  title     = {HiFi-GAN: Generative Adversarial Networks for Efficient and High Fidelity Speech Synthesis},
  booktitle = {Advances in Neural Information Processing Systems (NIPS)},
  year      = {2020},
  volume    = {33},
  pages     = {17022--17033}
}

@inproceedings{li2023freevc,
  author    = {J. Li and W. Tu and L. Xiao},
  title     = {Freevc: Towards high-quality text-free one-shot voice conversion},
  booktitle = {IEEE International Conference on Acoustics, Speech and Signal Processing (ICASSP)},
  year      = {2023},
  pages     = {1--5}
}

@inproceedings{baevski2020wav2vec,
  author    = {A. Baevski and Y. Zhou and A. Mohamed and M. Auli},
  title     = {wav2vec 2.0: A Framework for Self-Supervised Learning of Speech Representations},
  booktitle = {Advances in Neural Information Processing Systems (NIPS)},
  editor    = {H. Larochelle and M. Ranzato and R. Hadsell and M. Balcan and H. Lin},
  volume    = {33},
  year      = {2020},
  publisher = {Curran Associates, Inc.},
  pages     = {12449--12460}
}

@article{zhang2021initial,
  title={An initial investigation for detecting partially spoofed audio},
  author={Zhang, Lin and Wang, Xin and Cooper, Erica and Yamagishi, Junichi and Patino, Jose and Evans, Nicholas},
  journal={arXiv preprint arXiv:2104.02518},
  year={2021}
}

@article{Zhang2021Arxiv,
  author    = {L. Zhang and X. Wang and E. Cooper and J. Yamagishi},
  title     = {Multi-task learning in utterance-level and segmental-level spoof detection},
  journal   = {arXiv preprint arXiv:2107.14132},
  year      = {2021}
}

@article{PartialSpoof2022,
  author    = {L. Zhang and X. Wang and E. Cooper and N. Evans and J. Yamagishi},
  title     = {The partialspoof database and countermeasures for the detection of short fake speech segments embedded in an utterance},
  journal   = {IEEE/ACM Transactions on Audio, Speech, and Language Processing},
  volume    = {31},
  pages     = {813--825},
  year      = {2022}
}

@inproceedings{chan2022speechsplit,
  author = {Chak Ho Chan and Kaizhi Qian and Yang Zhang and Mark Hasegawa-Johnson},
  title = {Speechsplit2.0: Unsupervised speech disentanglement for voice conversion without tuning autoencoder bottlenecks},
  booktitle = {IEEE International Conference on Acoustics, Speech and Signal Processing (ICASSP)},
  year = {2022},
  pages = {6332--6336}
}

@inproceedings{chen2021againvc,
  author = {Y. Chen and D. Wu and T. Wu and H. Lee},
  title = {Again-vc: A one-shot voice conversion using activation guidance and adaptive instance normalization},
  booktitle = {IEEE International Conference on Acoustics, Speech and Signal Processing (ICASSP)},
  year = {2021},
  pages = {5954--5958}
}

@inproceedings{tang2022avqvc,
  author = {Huaizhen Tang and Xulong Zhang and Jianzong Wang and Ning Cheng and Jing Xiao},
  title = {Avqvc: One-shot voice conversion by vector quantization with applying contrastive learning},
  booktitle = {IEEE International Conference on Acoustics, Speech and Signal Processing (ICASSP)},
  year = {2022},
  pages = {4613--4617}
}

@inproceedings{cai2023waveform,
  title={Waveform boundary detection for partially spoofed audio},
  author={Cai, Zexin and Wang, Weiqing and Li, Ming},
  booktitle={IEEE International Conference on Acoustics, Speech and Signal Processing (ICASSP)},
  pages={1--5},
  year={2023},
  organization={IEEE}
}

@inproceedings{xie2024efficient,
  title={An Efficient Temporary Deepfake Location Approach Based Embeddings for Partially Spoofed Audio Detection},
  author={Xie, Yuankun and Cheng, Haonan and Wang, Yutian and Ye, Long},
  booktitle={IEEE International Conference on Acoustics, Speech and Signal Processing (ICASSP)},
  pages={966--970},
  year={2024},
  organization={IEEE}
}

@article{zhong2024enhancing,
  title={Enhancing partially spoofed audio localization with boundary-aware attention mechanism},
  author={Zhong, Jiafeng and Li, Bin and Yi, Jiangyan},
  journal={arXiv preprint arXiv:2407.21611},
  year={2024}
}

@inproceedings{radford2023robust,
  title={Robust speech recognition via large-scale weak supervision},
  author={Radford, Alec and Kim, Jong Wook and Xu, Tao and Brockman, Greg and McLeavey, Christine and Sutskever, Ilya},
  booktitle={International conference on machine learning (ICML)},
  pages={28492--28518},
  year={2023},
  organization={PMLR}
}

@article{yi2021half,
  title={Half-truth: A partially fake audio detection dataset},
  author={Yi, Jiangyan and Bai, Ye and Tao, Jianhua and Ma, Haoxin and Tian, Zhengkun and Wang, Chenglong and Wang, Tao and Fu, Ruibo},
  journal={arXiv preprint arXiv:2104.03617},
  year={2021}
}

@article{yadav2020end,
  title={End-to-end named entity recognition from english speech},
  author={Yadav, Hemant and Ghosh, Sreyan and Yu, Yi and Shah, Rajiv Ratn},
  journal={arXiv preprint arXiv:2005.11184},
  year={2020}
}

@inproceedings{tak2021end,
  title={End-to-end anti-spoofing with rawnet2},
  author={Tak, Hemlata and Patino, Jose and Todisco, Massimiliano and Nautsch, Andreas and Evans, Nicholas and Larcher, Anthony},
  booktitle={IEEE International Conference on Acoustics, Speech and Signal Processing (ICASSP)},
  pages={6369--6373},
  year={2021},
  organization={IEEE}
}

@inproceedings{jung2022aasist,
  title={Aasist: Audio anti-spoofing using integrated spectro-temporal graph attention networks},
  author={Jung, Jee-weon and Heo, Hee-Soo and Tak, Hemlata and Shim, Hye-jin and Chung, Joon Son and Lee, Bong-Jin and Yu, Ha-Jin and Evans, Nicholas},
  booktitle={IEEE international conference on acoustics, speech and signal processing (ICASSP)},
  pages={6367--6371},
  year={2022},
  organization={IEEE}
}

@inproceedings{kang2023grad,
  title={Grad-stylespeech: Any-speaker adaptive text-to-speech synthesis with diffusion models},
  author={Kang, Minki and Min, Dongchan and Hwang, Sung Ju},
  booktitle={IEEE International Conference on Acoustics, Speech and Signal Processing (ICASSP)},
  pages={1--5},
  year={2023},
  organization={IEEE}
}

@inproceedings{wu2025diffcss,
  title={DiffCSS: Diverse and Expressive Conversational Speech Synthesis with Diffusion Models},
  author={Wu, Weihao and Lin, Zhiwei and Zhou, Yixuan and Li, Jingbei and Niu, Rui and Wu, Qinghua and Cao, Songjun and Ma, Long and Wu, Zhiyong},
  booktitle={IEEE International Conference on Acoustics, Speech and Signal Processing (ICASSP)},
  pages={1--5},
  year={2025},
  organization={IEEE}
}

@inproceedings{han2025stable,
  title={Stable-TTS: Stable Speaker-Adaptive Text-to-Speech Synthesis via Prosody Prompting},
  author={Han, Wooseok and Kang, Minki and Kim, Changhun and Yang, Eunho},
  booktitle={IEEE International Conference on Acoustics, Speech and Signal Processing (ICASSP)},
  pages={1--5},
  year={2025},
  organization={IEEE}
}

@inproceedings{zuo2025enhancing,
  title={Enhancing expressive voice conversion with discrete pitch-conditioned flow matching model},
  author={Zuo, Jialong and Ji, Shengpeng and Fang, Minghui and Jiang, Ziyue and Cheng, Xize and Yang, Qian and Liu, Wenrui and Zhang, Guangyan and Tu, Zehai and Guo, Yiwen and others},
  booktitle={IEEE International Conference on Acoustics, Speech and Signal Processing (ICASSP)},
  pages={1--5},
  year={2025},
  organization={IEEE}
}

@article{vaswani2017attention,
  title={Attention is all you need},
  author={Vaswani, Ashish and Shazeer, Noam and Parmar, Niki and Uszkoreit, Jakob and Jones, Llion and Gomez, Aidan N and Kaiser, {\L}ukasz and Polosukhin, Illia},
  journal={Advances in neural information processing systems (NIPS)},
  volume={30},
  year={2017}
}

@article{van2016wavenet,
  title={Wavenet: A generative model for raw audio},
  author={Van Den Oord, Aaron and Dieleman, Sander and Zen, Heiga and Simonyan, Karen and Vinyals, Oriol and Graves, Alex and Kalchbrenner, Nal and Senior, Andrew and Kavukcuoglu, Koray and others},
  journal={arXiv preprint arXiv:1609.03499},
  volume={12},
  pages={1},
  year={2016}
}
\thispagestyle{empty}
\end{document}